# Evaluation of Feature Detector-Descriptor for Real Object Matching under Various Conditions of Ilumination and Affine Transformation

Novanto Yudistira[1], Achmad Ridok[2], Moch Ali Fauzi[3]

1) Yudistira Pictures; Universitas Brawijaya cbasemaster@gmail.com; yudistira@ub.ac.id
2) Universitas Brawijaya acridokb@ub.ac.id
3) Universitas Brawijaya moch.ali.fauzi@ub.ac.id

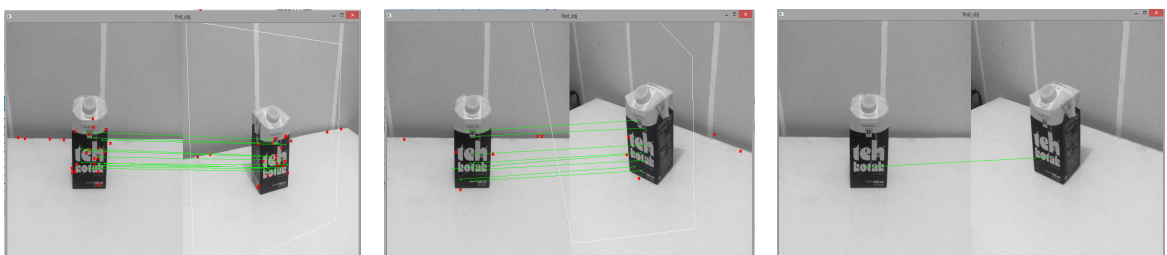

**Figure 1** Evaluation of feature repeatability between two identical objects with different camera viewpoint

## ABSTRACT

*This study attempts to provide explanations, descriptions and evaluations of some most popular and current combinations of description and descriptor frameworks, namely SIFT, SURF, MSER, and BRISK for keypoint extractors and SIFT, SURF, BRISK, and FREAK for descriptors. Evaluations are made based on the number of matches of keypoints and repeatability in various image variations. It is used as the main parameter to assess how well combinations of algorithms are in matching objects with different variations. There are many papers that describe the comparison of detection and description features to detect objects in images under various conditions, but the combination of algorithms attached to them has not been much discussed. The problem domain is limited to different illumination levels and affine transformations from different perspectives. To evaluate the robustness of all combinations of algorithms, we use a stereo image matching case.*

## 1. Introduction

There are several pattern recognition approaches by means of learning, structural matching and template matching. Pattern recognition with learning uses the statistical approach to model the right pattern to be able to make decisions. Specifically for template matching, it does pairing similar objects, which is usually represented by pixels as feature representation. On computer vision, representation of features should be as unique as possible to be beneficial for the task of discrimination. One method for obtaining feature representation is through a detector-descriptor method. It is known to be useful to support a related video processing application either an approximation or object tracking. Several studies of this method have been conducted using various descriptors with a suitable approach. Mikolajczyk et al, 2005 proposes a new way to evaluate feature extraction and description by creating a framework that provides a set of repeatability test from controlled projection positions and photometric transformations. Several evaluations have been made in this study by using detector-descriptor combinations of more than 100 objects viewed from the varied viewpoints and 144 images calibrated under various conditions. The results of its simple evaluation have shown to be robust for classification depending on the dataset's characteristics. In addition, this study has tested with a variety of 64D, 96D and 128D scales of SIFT (Scale Invariant Feature Transform) that showed good performance despite having large computing costs. It has been noted that SURF (Speeded Up Robust Features) also gives good classification results. In a study conducted Bay et al, 2008, they has proposed a fully SURF for image matching with affine invariant with accurate and efficient results. Lowe et al, 2009 proposes ASIFT for affine transformation cases that can handle tilt transitions up to 36 degrees and more. Alahi et al, 2012 has evaluated local features and kernels for texture classification and object categories. While Donoser et al., 2006 reported that a combination of Gaussian Difference (DoG) or MSER and SIFT detector features with DAISY descriptor provides the best combination of matching objects for a given dataset.

Mikolajczyk et al, 2005, have performed some detector and descriptor evaluations. However, evaluation of image matching with stereo objects along with the presentation of evaluation using various combinations of detector-descriptors with more additional results still have to be explored. Therefore this study was conducted.

The systematic nature of this paper is structured as follows: first, the introductory section which discuss brief descriptions of why this study was conducted as well as a review of related research. Second, related works which bring bright sight of state of the art's detector and descriptor methods. Third, we explain about our general evaluation setup and fourth, result and discussion.

## 2. Related Works
### 2.1. Detector Features

Some feature extractors and feature descriptors have recently been reported to provide good results for object matching. Some extractor-descriptor combinations can be generated and tested under different image conditions. As stated by Alahi et al, 2012, matching results highly depend on the combination of detectors. Each extractor or descriptor has unique characteristics. For example, binary based features such as BRISK (Binary Robust Invariant Scalable Points) act as extractors and descriptions as well. It is reported that BRISK can provide good quality in matching with little operational time. The other extractor is SIFT which is an extractor using maxima and minima of DoG in scaled space.

For feature detectors, researchers use MSER (Maximally Stable Extremal Region), SIFT, SURF, BRISK in their resilience considerations to discover potential feature areas.

### 2.2. Image Feature Detectors

Extracted features must be different but produce many keypoints. For this purpose a strong matching algorithm is used so that the results are not affected by conditions such as affine distortion, change of point of view, rotation and lighting. The more keypoints are obtained, it will have high possibility to produce a better object matching.

The first method we investigate is SIFT which aims to detect invariant objects on scale, rotation or representation by exploring the extreme local detection space to collect object features with DoG differences (equation 1). The scale of keypoint space is found by a sampling pattern consisting of dots by processing a gradient of local intensity.

$$D(x,y,\sigma) = (G(x,y,k\sigma)-G(x,y,\sigma))*I(x,y) \quad (1)$$

where

$$G(x,y,\sigma) = \frac{1}{2\pi\sigma^2} e^{-(x^2+y^2)/2\sigma^2}$$

The SIFT method is done by creating a pyramid of filter convolution and creating a scale space. The size of the scale space depends on the size of the image. Lowe suggests the use of 4 octaves using DoG (Gaussian differences) in the various dimensions (Lowe et al., 2009). While to accommodate invariant rotation, SIFT uses orientation gradients (equation 2) and the value is given at each keypoint.

$$m(x,y) = \sqrt{(L(x+1,y)-L(x-1,y))^2 + (L(x,y+1)-L(x,y-1))^2} \quad (2)$$
$$\theta(x,y) = \tan^{-1}((L(x,y+1)-L(x,y-1))/(L(x+1,y)-L(x-1,y)))$$

SURF uses a Hessian-based matrix (equation 3) where the determinant is used to find the desired area to be a candidate for a keypoint. Technically, images are inseparable from convolution filters, which are a box. Keypoints can be detected using a determinant matrix that maximizes / minimizes the convolution pyramid. While to accommodate the invariant scale, can be used in various scales.

$$\det(H_{approx}) = D_{xx}D_{yy} - (wD_{xy})^2 \quad (3)$$

Another detector, MSER, is used to detect a curvature-scale space. Imagine if we have a set of thresholding values, the image can change the white region to black and vice versa depending on thresholding increases or decreases. The most stable thresholding rate is determined as a representation area. Using an elliptical shape, keypoints are detected. MSER only accommodates keypoint detection. The recent modification of MSER is to move towards a more stable level of affine transformation, scale, lighting and invariant rotation.

### 2.3. Image Feature Descriptors

The characteristics of each keypoint can be illustrated by the description. Each of these descriptions should be unique so that it can be used for matching or dependent training. SIFT uses pixel samples from the pyramid level where the keypoint is detected. Furthermore, from some keypoint, it is described with a gradient orientation that represents the corresponding gradient.

FREAK is a keypoint descriptors inspired by human vision, particularly the retina. Alahi et al, 2012 explains that FREAK is generally faster in calculations with lower memory usage and is more efficient than its competitors such as SIFT, SURF or BRISK. Thus FREAK will remove the keypoint used by the BRISK descriptor and will reduce the calculation load.

SURF uses descriptor-based distribution. Each keypoint value is extracted with Haar Wavelet response for x and y directions. Next, the value of the descriptor is represented by a square window centered around the keypoint. SURF uses the number of Haar wavelets in each region as a vector feature using equation 4. This integrates gradient information in subpatches, which is more practical than the individual gradient orientation of SIFT.

$$v = (\sum d_x, \sum d_y, \sum |d_x|, \sum |d_y|) \qquad (4)$$

BRISK uses the following formulation in equation 5 to generate binary based descriptors.

$$g = \begin{pmatrix} g_x \\ g_y \end{pmatrix} = \frac{1}{L} \sum_{(p_i, p_j) \in G} g(p_i, p_j)$$
$$b = \begin{cases} 1, I(p_j^\alpha, \sigma_j) > I(p_i^\alpha, \sigma_i) \\ 0, otherwise \end{cases} \qquad (5)$$

### 3. Evaluation Setup

In general, the research flow diagram as shown in Figure 2. To evaluate the detector feature and description, some object references and test objects captured with mobile devices are used as data. Each ektsraktor - descriptor, each has different characteristics.

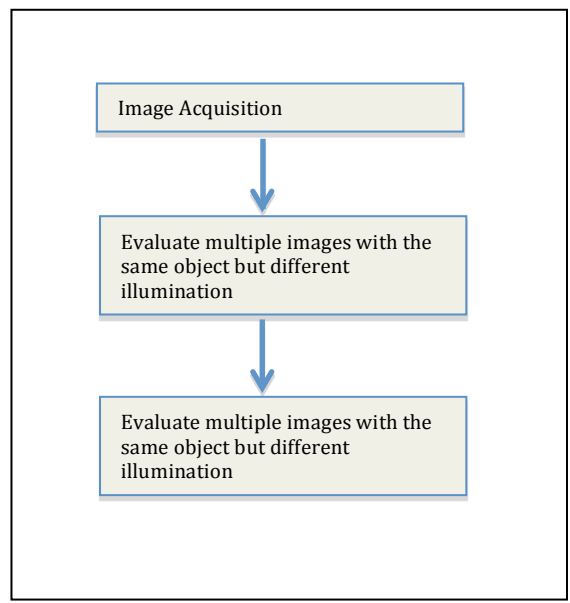

**Figure 2 Research flow diagram**

Object matching is evaluated using repeatability metric. Evaluation is done by how many keypoints are detected in accordance with the reference image. The evaluation results are determined by how well the object recognizes the image with the four basic transformations of illumination and affine transformation. The affine transformation is related to a change of view angle with some degree of projection angle. This is the main problem that attempted to be solved by determining the features of the detector - descriptor match.

### 3.1 Image Acquisition

The acquisition of photos is taken through mobile devices collected in different shapes that meet various illumination and affine transformations. The position of the photographed object is arranged so that it can accommodate all possible values. For example, in the affine transformation test, every 20-degree image tested matches an inappropriate reference image similar to this angle. This is done to try the detector - descriptors naturally so it is pretty close to in the case of real world applications.

### 3.2. Evaluation using multiple images with the same object but different illuminations

Illumination is related to the intensity of the image set using various exposures as shown in Figure 3. There are 4 exposure values to be tested: +4 (light enough), 7 (light), -4 (dark enough), and 7 (dark). While the image having an exposure value of 0 (normal) is set as a reference. This is done to accommodate all levels of illumination that may still be capable of being processed by image processing algorithms.

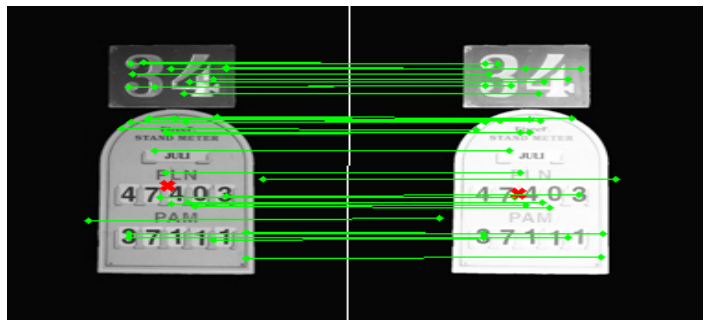

Figure 3  Matching two different illuminated images

### 3.3. Evaluate multiple images with the same object but different affine transformations

Evaluation of multiple images with the same object but different affine transformations is the most difficult test. This is because the image is tested based on different angles or perspectives as shown in Figure 4. Several trials are conducted by arranging different levels of view. Drag the test image taken every 20 degree projection.

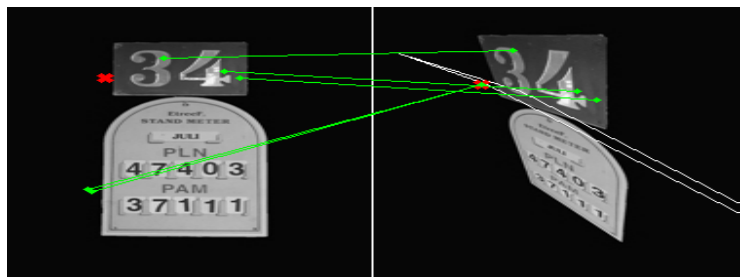

Figure 4  Matching two different images of affine transformation

This evaluation is similar to what Mikolajczyk et al. The evaluation is done by a combination of detector-descriptor test that refers to how many matches are right and wrong. In this case the feature matching algorithm (keypoint) is used kNN dengsan value k = 2. This means that the two closest keypoins of each image are tested with each keypoint in the reference image which will be a cluster containing two pairs. By giving a ratio of 0.75 then each cluster will detect only the remaining pair by filtering out another pair having a distance greater than 0.75 multiplied by the closest distance of both. It is necessary to filter multiple pairs to obtain precise results.

### 3.4. Evaluate multiple images with the same object but different affine transformations using ALOI dataset

The evaluation was conducted with ALOI (Amsterdam Library of Object Images) which is a collection of thousands of small objects and captured for scientific purposes. The dataset has several variations including the angle of sight (affine transformation), illumination angle, and color illumination on each captured object. In addition there are also objects with different angles of view with a wide angle (stereo images). Overall, each object has hundreds of images totaling 110,250 images.

In Figure 5, each object is recorded with only one out of five lights on, producing five different angle lighting (condition 13-15). By switching the camera, and turning the stage towards the camera, the lighting bow is almost rotated by 15 (camera c2) and 30 degrees (camera c3), respectively. Therefore, aspects of the object seen by each camera are identical, but the direction of light has shifted by 15 and 30 degrees in azimuth. In total, these results are in 15 different angles of illumination.

In addition, the combination of lights used to illuminate the object. Turning on two lights on the side of the object produces tilted illumination from right (condition 16) and left (condition 17). Turning on all lights (condition 18) results in a kind of hemispherical illumination, though limited to the narrower lighting sector of the right hemisphere. In this way, a total of 24 different lighting conditions are generated.

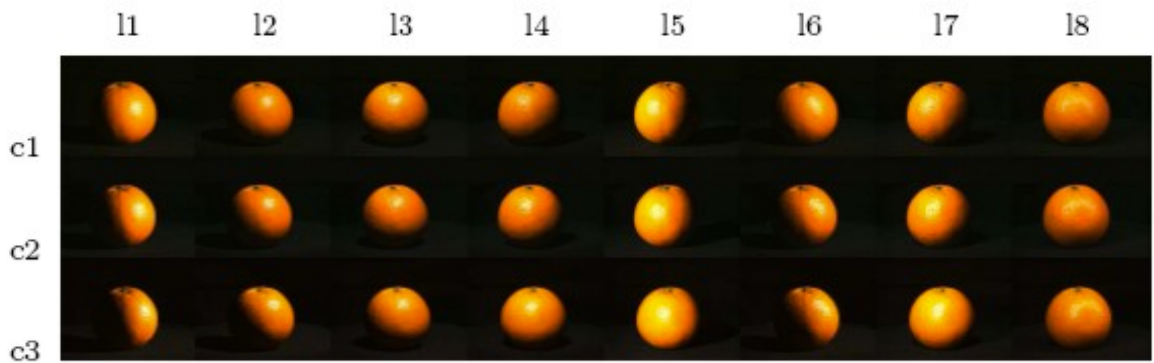

Figure 5  ALOI dataset with different combination of lights

Each object is recorded in the frontal view, with all five lights turned on. The color temperature of the lighting changed from 2175K to 3075K. The white balance is set at 3075K, so that the object is illuminated under the reddish illumination with a condition value of 110, 120, ..., 250.

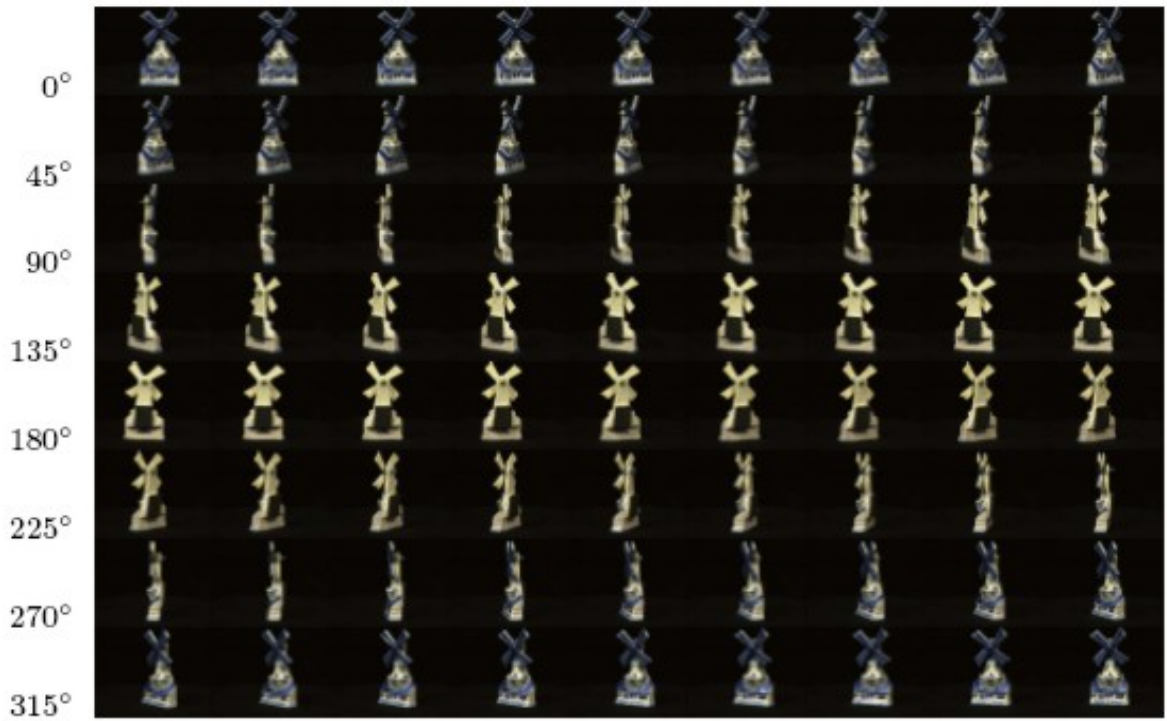

Figure 6  ALOI dataset with different rotation

In Figure 6, small rotations starting from respective angles are provoded. In Figure 7, by rotating the object, the central image (c configuration), the right image (r configuration) and the left image (configuration l) can be made. The middle-left and middle-right combinations produce two pairs of 15 degrees of stereo base, while the left-right pair combination produces 30 degrees of base pair of stereo.

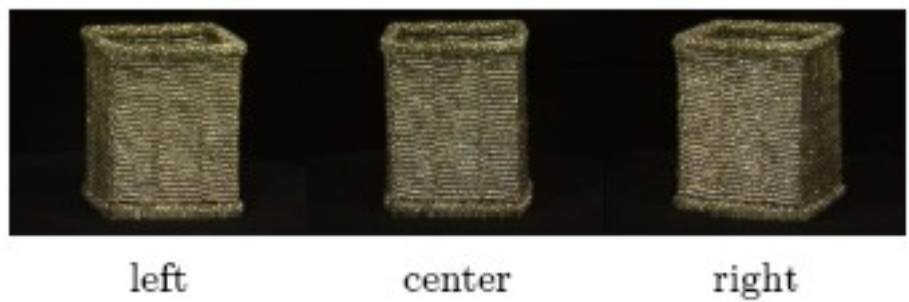

Figure 7  ALOI dataset of 3 angle configurations

## 4.  Results and Discussion
### 4.1. Illumination Differences

In the example matching two objects, shown by example between scissors with different illumination. The test results using SIFT as a detector and SURF as descriptors in Figure 8 show that there are several matching keypoints between two different scissor objects whose exposure value is -4 and +4. However, between the two matching keypoints there are some incorrect pairs.

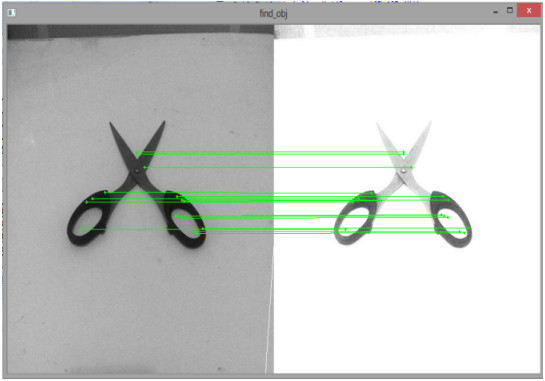

Figure 8  SIFT-SURF with exposure -4

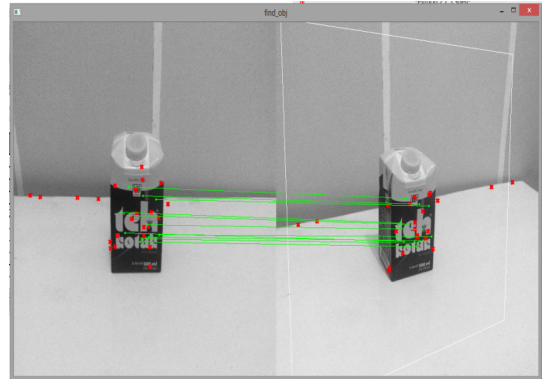

Figure 9  SIFT-SURF with affine transformation (viewpoint of 20° to left)

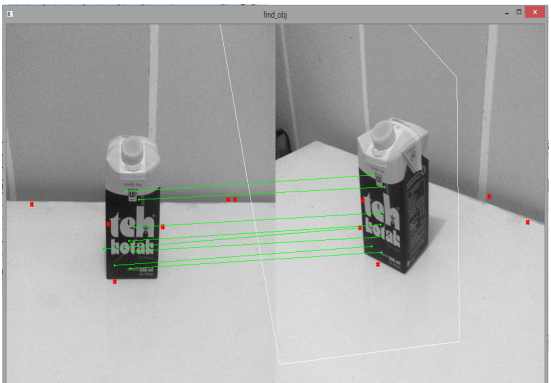

Figure 10  SURF-BRISK with affine transform (viewpoint of 20° to right)

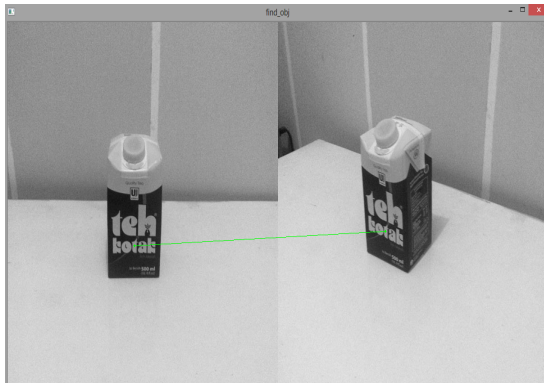

Figure 11  MSER-BRISK with affine transform (viewpoint of 20° to right)

### 4.2. Affine Transformation Differences

Matching two objects is shown by example between tea packing boxes with different combinations of different affine transformation descriptors. In Figure 9 shows the test results using a combination of SIFT as a detector and SURF as descriptors under affine transforms of different view angles 20 degrees to the left on the camera sensor side, there are some correct matching keypoints. In Figure 10 with SURF combination as a detector and BRISK as descriptors under affine transform conditions different angles of 20 degrees to the right appear to have some correct matching keypoint. In Figure 11 with the combination of MSER as a detector and BRISK as a descriptor in a different affine transform state of view of 20 degrees to the right seen there are some correct matching keypoint. Although many keypoints look right but we have to count again in detail what is right and what is wrong, what is the cause, or by using a clustering technique to determine which keypoint fits correctly and what is wrong.

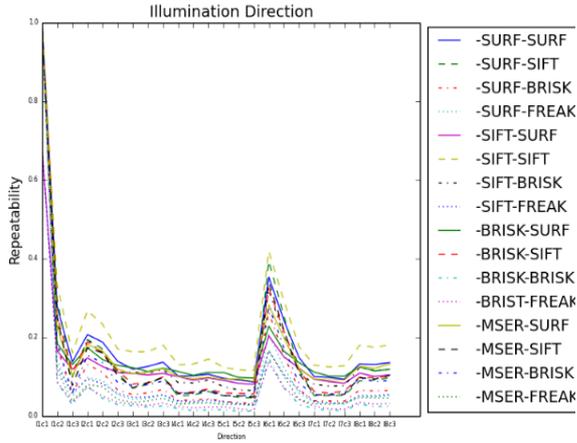

Figure 12 Repeatability on various illumination directions

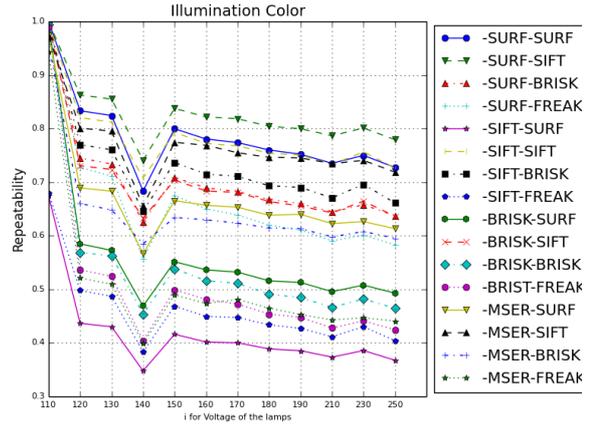

Figure 13 Repeatability on various illumination colors

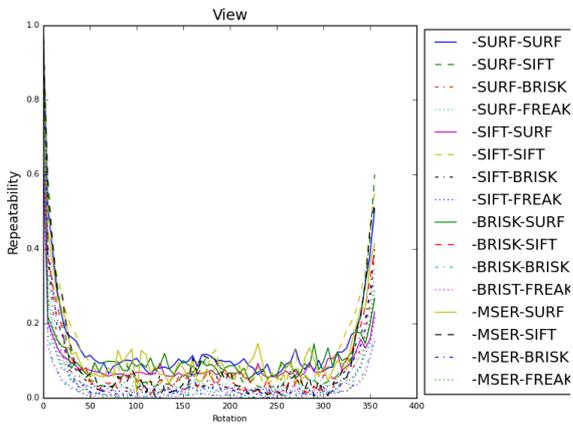

Figure 14 Repeatability on various illumination views

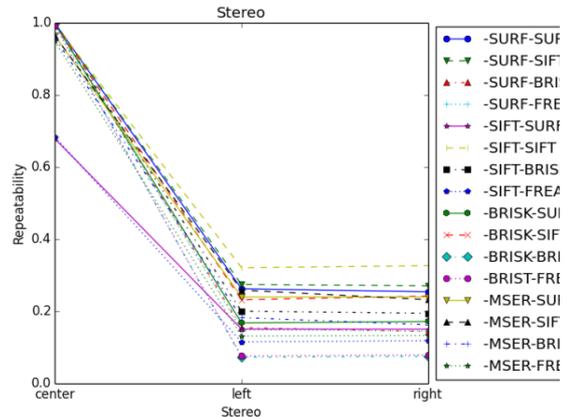

Figure 15 Repeatability on various stereo illumination

### 4.3. ALOI Dataset

To test the difference illumination and affine transformation, ALOI dataset (J.M. Gaosebroek et al., 2005) is used as a part of experiment due to its large number of samples. Each part of tests include three resolutions that are full resolution, half resolution and quarter resolution. The first test is based on variations of illumination from various directions. It does directional illumination test by providing variations in combination of camera angle and flame of light. This combination produces 24 different variations. As shown in Figure 12, the combination of SIFT - SIFT provides the best repeatability for directional illumination cases. In addition, the SURF-SIFT and SURF-SURF methods also provide fairly good repeatability results and are not much different from the SIFT-SIFT method. For color illumination, as shown in Figure 13, the combination of SURF - SIFT provides the best repeatability for color illumination cases. In addition, the SURF-SURF and SIFT-SIFT methods also provide fairly good repeatability results and are not much different from the SURF-SIFT method. For illumination views, as shown in Figure 14, the combination of SURF - SURF provides the best repeatability for view cases in full resolution images. While there are some other methods also repeatability results are not much different from the SURF-SURF method. For stereo illumination, As shown in Figure 15, the combination of SIFT - SIFT provides the best repeatability for stereo cases. In addition, the SURF-SIFT and SURF-SURF methods also provide fairly good repeatability results and are not much different from the SIFT-SIFT method. For Scaling, as shown in Figure 16, the combination of SIFT - SIFT provides the best repeatability for this case. In addition, the SURF-SURF and MSER-SURF methods also provide fairly good repeatability results and are not much different from the SIFT-SIFT method. As for the rotation, as shown in Figure 17, the combination of SIFT - SIFT provides the best repeatability for this case.

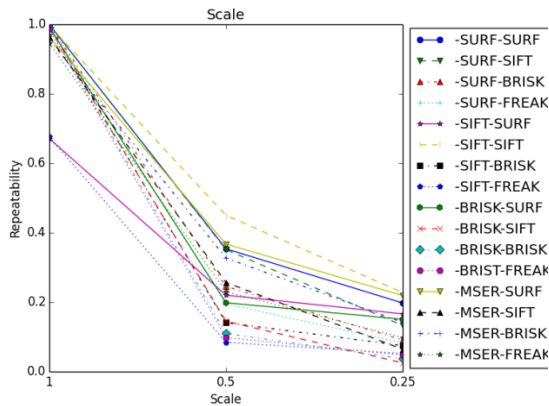
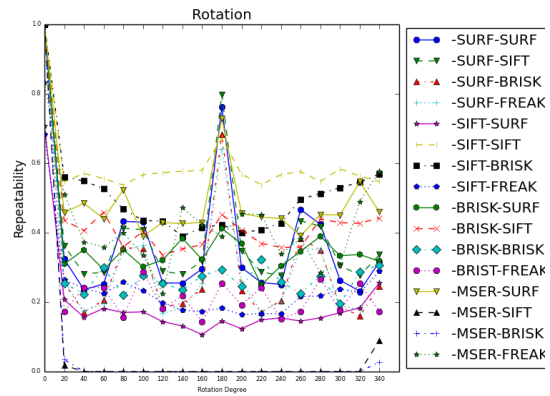

Figure 16 Repeatability on scaling      Figure 17 Repeatability in rotation

## 5. Conclusion

The combination of detectors and descriptors has been tested based on their keypoint repeatability. With only 7 objects tested with various illumination conditions and affinity transformations, it has been observed how detector and descriptor capability overcomes different illumination and affine transformation problems although quantification and analysis with statistics is required to reach a more conclusions. Furthermore, in addition to quantifying the number of matching and correct keypoints on each detector-descriptor combination, we will also solve the problem of how both matching points or keypoints can be considered exactly correct. We will also try out on large datasets with more diverse objects. The results of the test with this general dataset are expected to conclude whether the results of this evaluation will find a powerful combination of detectors and descriptors for the problem of illumination differences and affine transformations comparable to previous studies.